\documentclass[twocolumn]{article}
\usepackage{graphicx} 
\usepackage{float}
\usepackage{xcolor}
\usepackage{caption}
\usepackage{subcaption}
\usepackage{mwe}
\usepackage{amsmath,amssymb}
\usepackage{relsize}
\usepackage{array,booktabs}
\usepackage[round]{natbib}
\usepackage{url}
\renewcommand{\cite}[1]{\citep{#1}}
\usepackage{comment}
\newcolumntype{C}{>{$\displaystyle}c<{$}}
\newcommand\todo[1]{\textcolor{red}{Todo: #1}}
\newcommand\rigid[1]{\bar{#1}}
\newcommand\deformable[1]{\Tilde{#1}}

\providecommand{\keywords}[1]
{
  \small	
  \textbf{\textit{Keywords---}} #1
}

\title{Tracking Mouse\\from Incomplete Body-Part Observations and\\Deep-Learned Deformable-Mouse Model\\Motion-Track Constraint for Behavior Analysis\\[0.9em]\smaller{}The only thing about me is the way I walk}
\author{Olaf Hellwich\textsuperscript{1,6*}, Niek Andresen\textsuperscript{1,4,6},
Katharina Hohlbaum\textsuperscript{2,6}, Marc Boon\textsuperscript{3,6}, \\
Monika Kwiatkowski\textsuperscript{1},
Simon Matern\textsuperscript{1}, Patrik Reiske\textsuperscript{1,6},
Henning Sprekeler\textsuperscript{3,6}, \\ Christa Thöne-Reineke\textsuperscript{4,6},
Lars Lewejohann\textsuperscript{2,4,6}, Huma Ghani Zada\textsuperscript{1,5}, \\ 
Michael Brück\textsuperscript{6}, Soledad Traverso\textsuperscript{3,1,6}
\medskip
\\ \textsuperscript{1}TU Berlin, Computer Vision \& Remote Sensing
\\ \textsuperscript{2}German Centre for the
Protection of Laboratory Animals (Bf3R), \\ German Federal Institute for Risk Assessment (BfR)
\\ \textsuperscript{3}TU Berlin, Modeling of Cognitive Processes
\\ \textsuperscript{4}FU Berlin, Institute of Animal Welfare, Animal Behavior \\ and Laboratory Animal Science
\\ \textsuperscript{5}TU Berlin, Remote Sensing Image Analysis
\\ \textsuperscript{6}TU Berlin, Science of Intelligence Excellence Cluster
\\ \textsuperscript{*}corresponding author}
\date{September 23, 2024}

\begin{document}

\twocolumn[\vspace*{-1cm}
  \begin{@twocolumnfalse}

\maketitle

\begin{abstract}

Tracking mouse body parts in video is often incomplete due to occlusions such that - e.g. - subsequent action and behavior analysis is impeded. In this conceptual work, videos from several perspectives are integrated via global exterior camera orientation; body part positions are estimated by 3D triangulation and bundle adjustment. Consistency of overall 3D track reconstruction is achieved by introduction of a 3D mouse model, deep-learned body part movements, and global motion-track smoothness constraint. The resulting 3D body and body part track estimates are substantially more complete than the original single-frame-based body part detection, therefore, allowing improved animal behavior analysis.$^{1}$

\end{abstract}


\vspace{5mm}
  \end{@twocolumnfalse}
]

\section{Introduction}
\label{sec:intro}

\begin{figure*}[!tb]
    \centering
    \begin{subfigure}[b]{0.475\textwidth} 
        \centering
        \includegraphics[width=\textwidth]{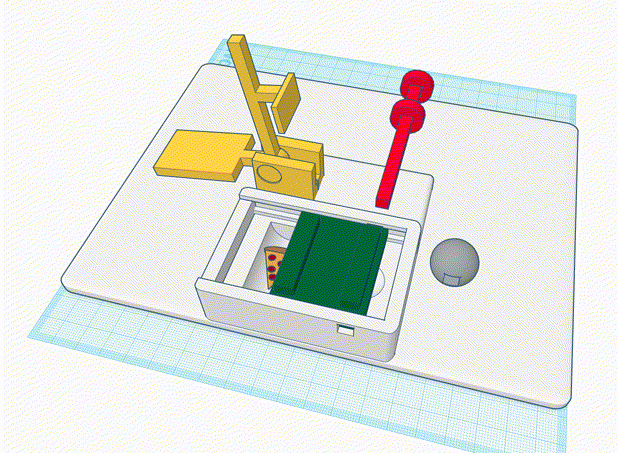}
        \caption[]%
        {{\small mouse lock box with lever (yellow), splint (red), ball (gray), and slider (green)}}    
        \label{fig:lockBox}
    \end{subfigure}
    \hfill
    \begin{subfigure}[b]{0.475\textwidth}  
        \centering 
        \includegraphics[width=\textwidth]{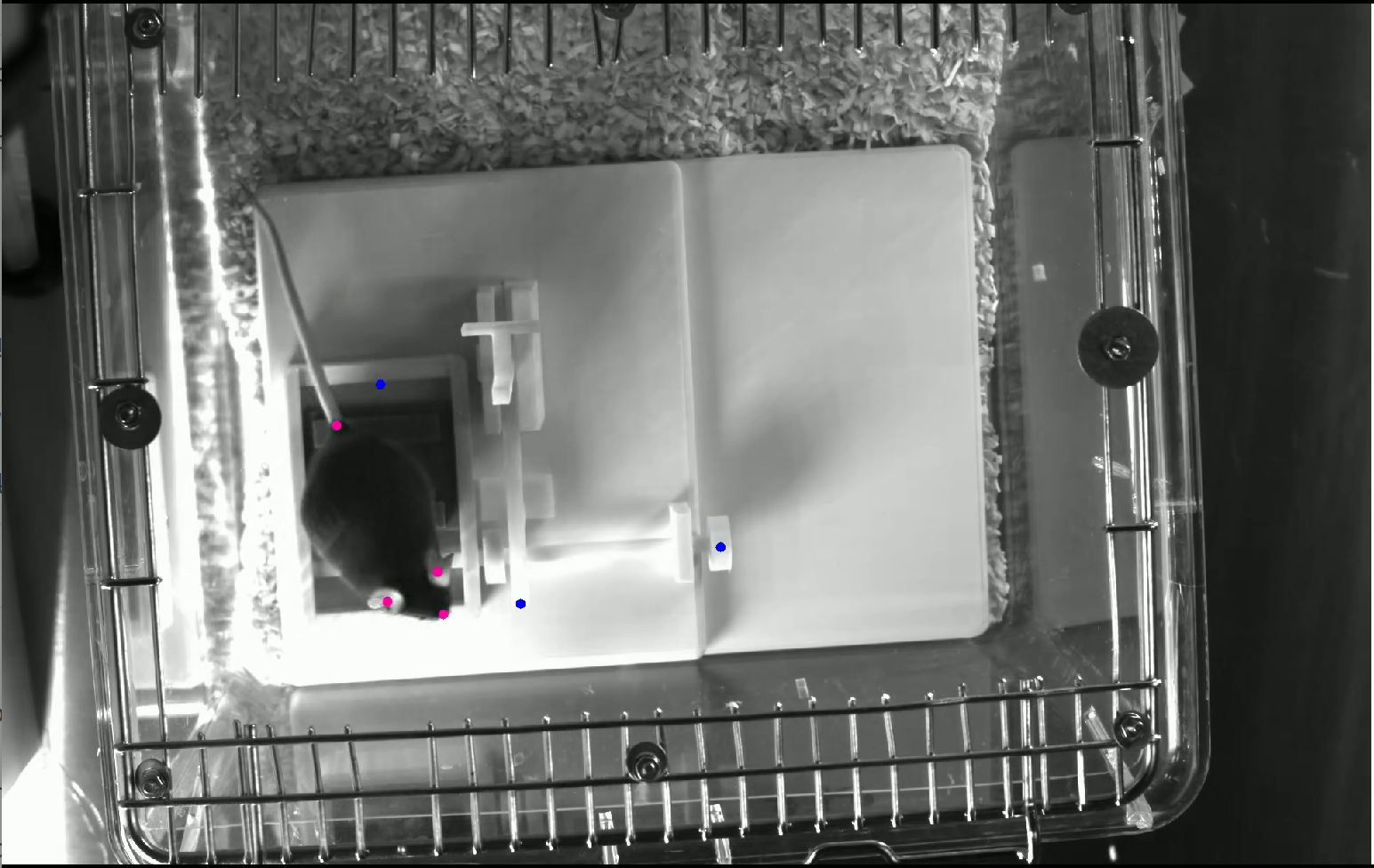}
        \caption[]%
        {{\small top view}}    
        \label{fig:topView}
    \end{subfigure}
    \vskip\baselineskip
    \begin{subfigure}[b]{0.475\textwidth}   
        \centering 
        \includegraphics[width=\textwidth]{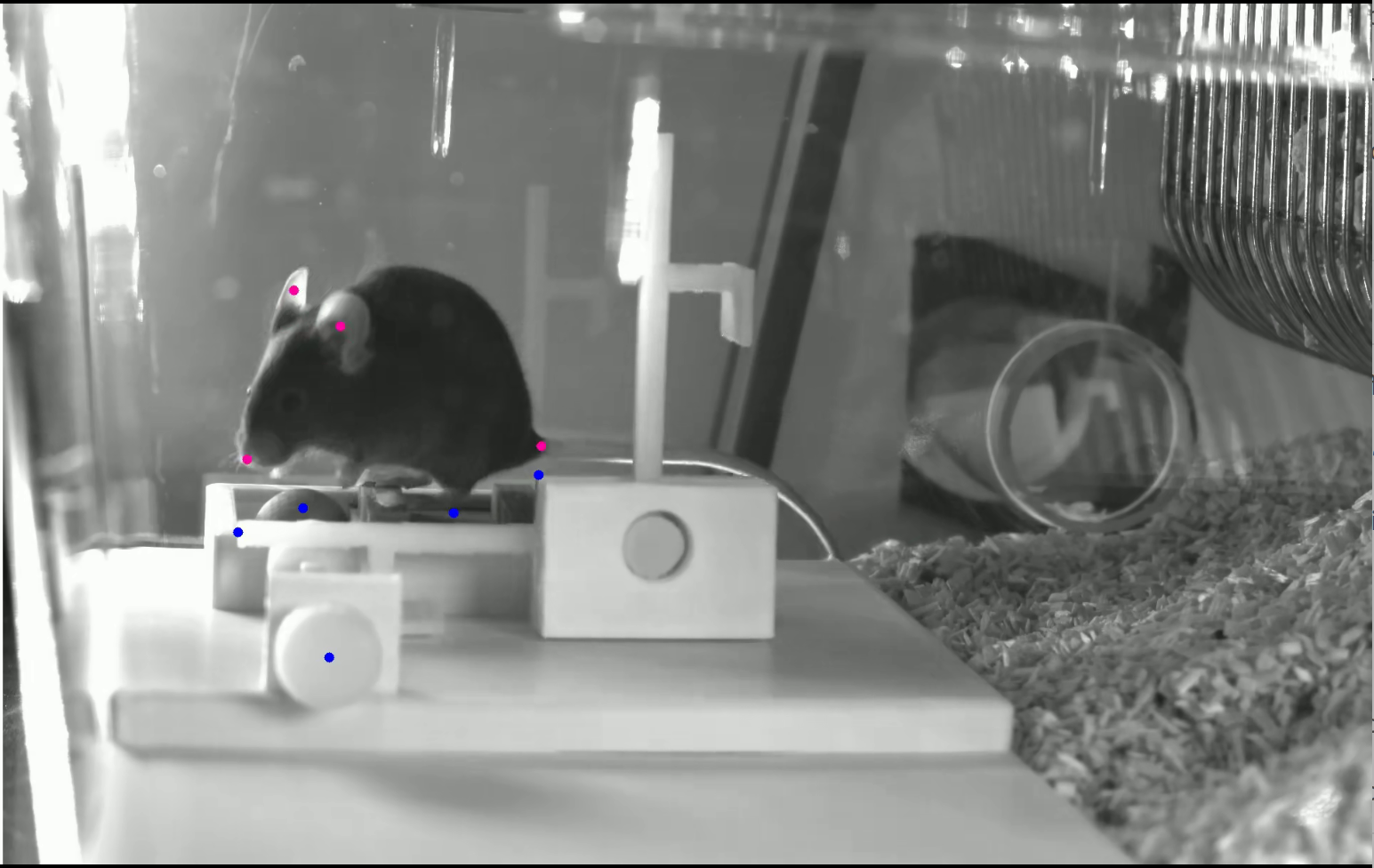}
        \caption[]%
        {{\small side view}}    
        \label{fig:sideView}
    \end{subfigure}
    \hfill
    \begin{subfigure}[b]{0.475\textwidth}   
        \centering 
        \includegraphics[width=\textwidth]{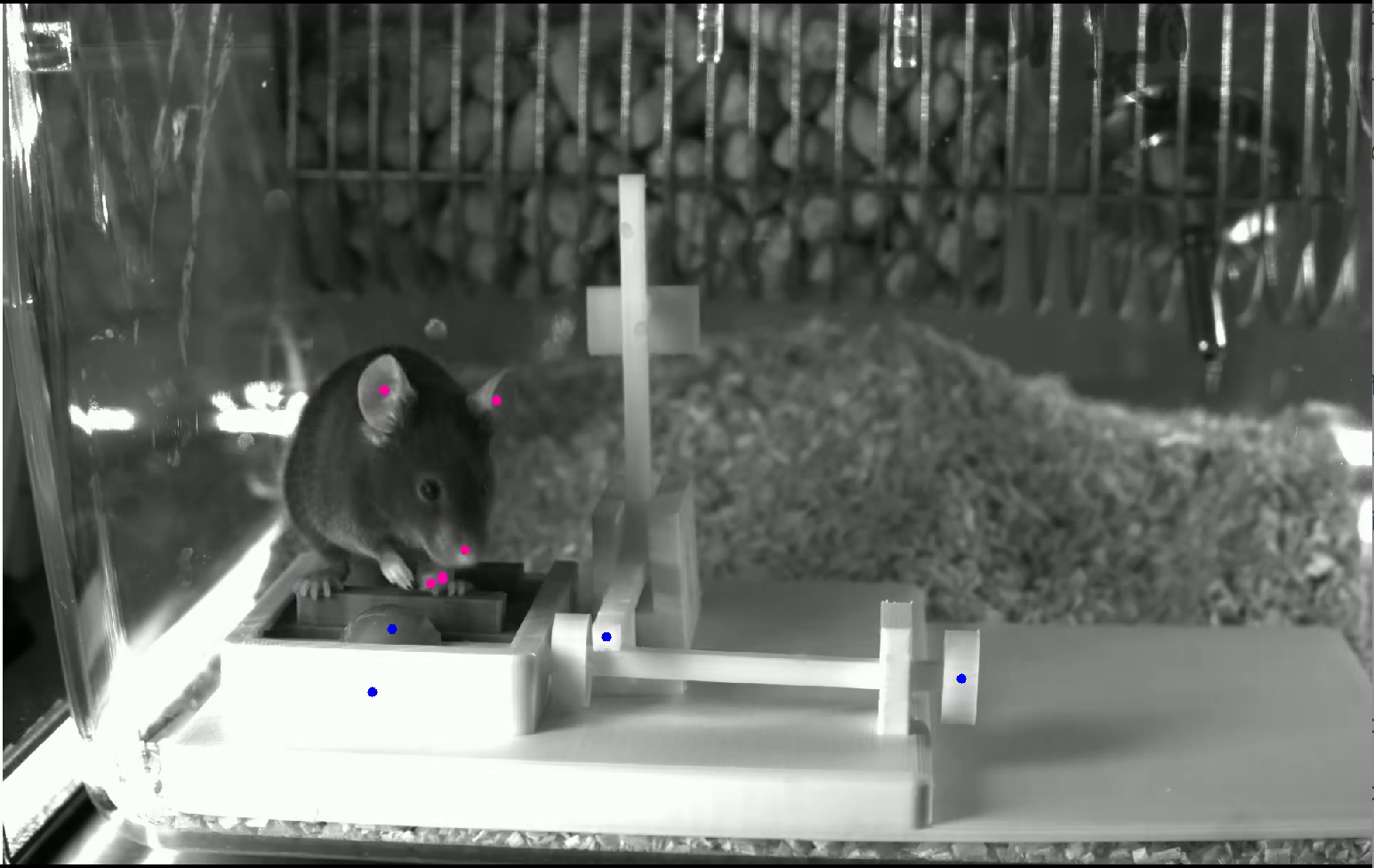}
        \caption[]%
        {{\small front view}}    
        \label{fig:frontView}
    \end{subfigure}
    \caption[]
    {Mouse lock box design and mouse in multi-cam video views solving the lock-box riddle} 
    \label{fig:lockBoxVideoViews}
\end{figure*}

In order to analyze how animals like mice solve problems, whether they are able to learn from previous experiences or from observing other members of their group, or whether they profit from exact reproduction of sequences of touch and application of force to objects or object parts, requires detailed observation of their movements and how they interact with the objects in their environment and other members of their group. In enriched natural environments such observations, usually done with help of video sensors, are not trivial (Fig.~\ref{fig:lockBoxVideoViews}). In previous work, as expected, we experienced substantial difficulties to observe how mice interact with the objects in their environment, whether, in which order, and in which way they touch object parts with paws, snout, or other parts of their bodies. Often the body parts essential for precisely understanding the action of the animal were subject to motion blur, occlusion by other parts of the mouse's body, other mice, or object's in the enriched cage. These problems can be reduced by the use of multiple video sensors observing the scenery in parallel.

However, it is obvious that more complex and costly set-ups of the sensor equipment only provides gradual relief, but does not solve the problem in general. Rather, and additionally, the available video sequences should be evaluated such that the required information is derived with the lowest-possible degree of uncertainty.

Depending on the final goal of the experiment, the required information to be derived from the video sequences is usually not the 3D positions where body parts are located as a function of time. For instance, in a mouse lock box experiment, as treated in \cite{Boon2024.07.29.605658}, the more important question is in which order a mouse operates the mechanical components of the lock box riddle (Fig.~\ref{fig:lockBox}) - e.g. first lever, then splint, then ball, finally slider lid, which is the only sequence of mechanisms that allows access to the hidden reward, an oats flake. Or, in a more advanced group observation setting, the question is whether the behavior of several mice indicates that the informed group members teach uninformed group members, how to successfully open the lock box.

While in principle it is justified to try to estimate more advanced motion or behavior categories directly from the video sequence, in this paper, as one alternative out of several, we advocate the use of a data evaluation process flow that provides 3D body part positions prior to make conclusions w.r.t.~behavior. First, the 3D body part positions are computed from multi-view images using a photogrammetric approach. Then according to how and where the body parts come closer to and get in touch with e.g. objects, conclusions w.r.t. behavior classes are drawn. In this paper, we introduce a concept for the first step, i.e. for 3D body part tracking from multi-view video where body part extraction from (single-view) video suffers substantially from occasional occlusions preventing immediate triangulation of 3D positions from body part extractions in synchronous multi-view video frames.

This paper is published on the occasion of the 60th birthday of Helmut Mayer.\footnote{This paper has been published in: Reinhardt, Wolfgang; Huang, Hai (editors): Festschrift f\"ur Prof. Dr.-Ing. Helmut Mayer zum 60. Geburtstag, Institut f\"ur Geod\"asie der Universit\"at der Bundeswehr M\"unchen, Vol. 101, 2024, pages 45 - 53; https://athene-forschung.unibw.de/150907} It is an early concept on things still to be done rather than a report on research results with an adequate analysis. In this respect it is unusual lacking scientific rigor by reporting results to be achieved by work not yet done. Having made this confession, we dare to advance the treatise congratulating the jubilee. A future follow-up version of this paper will eliminate this deficit.

\section{Previous Work}

The track constraint on the 3D motion of body parts introduced in Section \ref{sec:modelMethod} is inspired by approaches that 30 years ago led to the computation of 3D surface models from three-line-scanner cameras where camera paths were determined by orbit geometry or inertial measurements providing exterior orientations for line images \cite{ebner1993simulation, gruen2003sensor}. The inertial measurements are related to the sequences of line images by time stamps of both inertial measurement unit (IMU) and camera. The line images' IMU orientation data is interpolated for image acquisition times by e.g.~cubic spline interpolation. A similar approach has been taken in \cite{hellwich2000geocoding} for geocoding spaceborne Synthetic Aperture Radar interferograms including use of ground control points.

As mentioned in the introductory Section \ref{sec:intro}, in our previous work \cite{Boon2024.07.29.605658} multi-view video data is used to analyze how mice learn when solving lock-box problems. As to be inferred from Fig.~\ref{fig:lockBoxVideoViews}, three video cameras observe the animals from frontal, top and side view points, i.e.~with approximately 90° view direction difference.

\section{Model and Method}
\label{sec:modelMethod}

In this section, a mouse body model and a motion-track smoothness constraint is introduced in order to improve 3D body part tracking.

\subsection{Global Co-Ordinate System}

We assume that cameras as well as non-mouse objects are firmly mounted and do not move. 3D object points, e.g. lock box parts, are given in global co-ordinates. Camera orientations can e.g. be computed by spatial resection for each camera from observed lock box parts $\mathbf{x}_{i}^{k}$ according to
\begin{equation}
    \mathbf{x}_{i}^{k} = \mathbf{P}^{k} \mathbf{G}_i
\label{Eq:camOrient}
\end{equation}
with camera index $k$, projection matrices $\mathbf{P}^k$, and 3D lock box point in global co-ordinates $\mathbf{G}_i$. 

Decomposition of $\mathbf{P^k}$ provides the invertible rigid transformation $\mathbf{H}_g^k$ from global to 3D camera co-ordinates 
\begin{equation}
    \mathbf{P^k} = \mathbf{K}
\begin{bmatrix}
1 & 0 & 0 & 0\\
0 & 1 & 0 & 0\\
0 & 0 & 1 & 0\\
\end{bmatrix}
 \mathbf{H_g^k}
 \label{eq:intExtOrientDec}
\end{equation}
with $\mathbf{K}$ being the camera calibration matrix.

\subsection{Modelling the Relationship Between Body Parts and Camera Orientation}

Goal of the approach is to determine the track of the mouse's body and its parts in global 3D space.

However, tracks can also be determined in 3D mouse (model) co-ordinate system. Then, due to the motion of the mouse, the projection matrix $\mathbf{P}$ of the camera in 3D mouse model space is subject to change. We express this by introducing a time index $t$. So $\mathbf{P}_t^k$ refers to mouse model co-ordinate system.

For conceptual clarity, we note that $\mathbf{P}_t^k$ can be determined from the projections of three or more known 3D object points in a single video frame $I_t^k$, e.g. from three 3D mouse body parts, using the observation equations
\begin{equation}
    \label{Eq:observation}
    \mathbf{x}_{ti}^k = \mathbf{P}_t^k \mathbf{X}_i
\end{equation}
which are equivalent to Eq.~\ref{Eq:camOrient}.

As soon as less than three known body parts are observed, this is not possible any more. However, instead of body part observations a condition equation enforcing the continuity of $\mathbf{P}_t^k$ over time $t$ can be introduced. We call it motion-track constraint.

\subsection{Motion-Track Constraint}

The motion-track constraint is formulated as an interpolation of projection matrix $\mathbf{P}_{t}^k$ in-between neighboring projection matrices $\mathbf{P}_{t-2}^k$, $\mathbf{P}_{t-1}^k$, $\mathbf{P}_{t+1}^k$ and $\mathbf{P}_{t+2}^k$ using cubic spline interpolation. For instance, we suggest spline interpolation on all six exterior orientation parameters individually. For that purpose the $\mathbf{P}$ matrices are first decomposed according to \ref{eq:intExtOrientDec}. The resulting $\mathbf{H}_t^k$ providing transformations from mouse model co-ordinates to camera co-ordinates are further decomposed into rotation matrices $\mathbf{R}_t^k$ and translation vectors $\mathbf{t}_t^k$ according to
\begin{equation}
    \mathbf{H}_t^k = 
\begin{bmatrix}
\mathbf{R}_t^k & \mathbf{t}_t^k \\
\mathbf{0}^\text{T} & 1\\
\end{bmatrix}
\label{Eq:rigidTransDecomp}
\end{equation}
A transformation of the rotation matrix into a Rodriguez vector finalizes the decomposition of $\mathbf{H}_t^k$ into six parameters $\mathbf{p}_t^k$ of a rigid transformation constituting the mouse model state vector.

For each element $p_l$ of $\mathbf{p}_t^k$, for 5 time-sequential epochs a cubic spline interpolation $S(...)$ can be conducted for epoch $t$ and compared with the estimate $p_{l,t}$ resulting from the current video frame's body-part observations:
\begin{equation}
\label{Eq:spline}
    d = p_{l,t} - S(p_{l,t-2}, p_{l,t-1}, p_{l,t+1}, p_{l,t+2})
\end{equation}
$d$ is the difference between parameter $p_{l,t}$ and its spline-interpolated value $S(...)$ obtained from neighboring time epochs. I.e.~(\ref{Eq:spline}) constitutes the track constraint supporting smoothness.

The condition equations (\ref{Eq:spline}) may not be easy to handle as all six parameters $p_l$ of $\mathbf{p}_t^k$ are treated individually/independently. The better option is a comparison of the two transforms resulting from locally visible body parts and spline interpolation. So, the six spline-interpolated parameters are recombined to a transformation matrix $\mathbf{S}_t^k$. It is to be compared with $\mathbf{H}_t^k$.

\begin{figure}[htbp]
  \centering\includegraphics[width=0.5 \columnwidth]{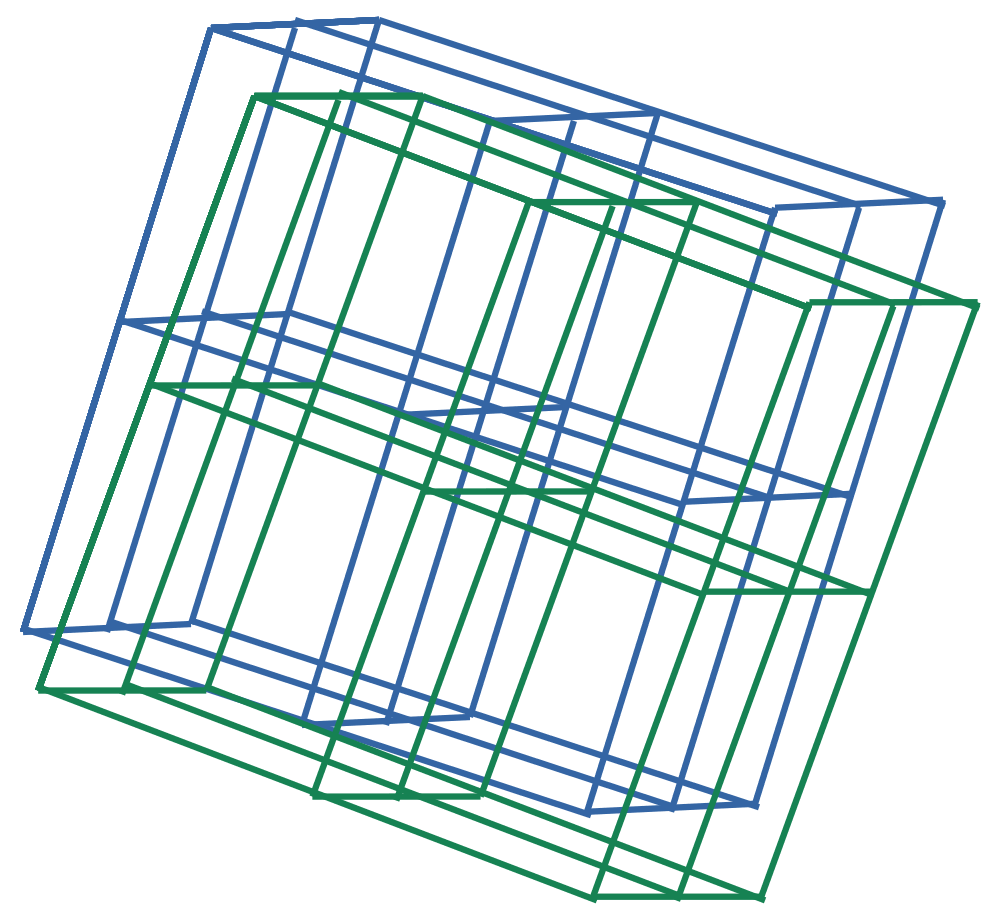}
  \caption{Grid before (blue) and after transform (green) with $\mathbf{H}_t^k {\mathbf{S}_t^k}^{-1}$}
  \label{fig:gridCompare}
\end{figure}

For the comparison of the two transforms, at the location of interest, i.e.~the center of gravity of the mouse, a 3D grid of at least $3^3$ grid positions covering the body of the mouse are defined. It is transformed by transform $\mathbf{H}_t^k {\mathbf{S}_t^k}^{-1}$. Then the RMSE is computed from all pairs of grid points - replacing the six differences according to (\ref{Eq:spline}). Fig.~\ref{fig:gridCompare} shows an example grid comparison.

\subsection{Rigid Mouse Model}
\label{sec:rigidMouse}

\begin{figure}[htbp]
  \centering\includegraphics[width=0.7 \columnwidth]{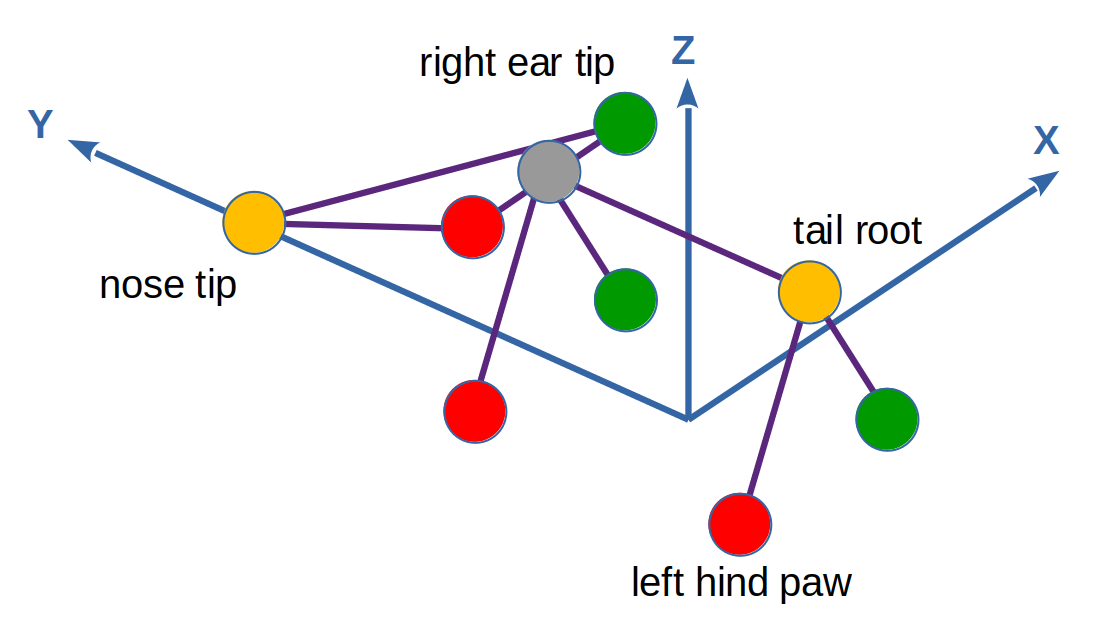}
  \caption{Rigid mouse model.}
  \label{fig:rigidMouseModel}
\end{figure}

The simplest mouse model assumption would be a rigid mouse (Fig. \ref{fig:rigidMouseModel}). In this case, the movements of body parts that are existing in reality would be interpreted as noise in the observations of image co-ordinates of these body parts. Then, despite un-modeled movements, all estimations can be conducted as formulated here.

Table \ref{tab:rigidMouseModel} shows rigid mouse model co-ordinates that are biologically reasonable.

\begin{table}[htbp]
\caption{Biologically realistic rigid mouse model co-ordinates.}
\centering
\begin{tabular}{ |c|c|c|c|c|c|c| } 
\hline
 body part & $X$ & $Y$ & $Z$ \\
 & [mm] & [mm] & [mm] \\
 \hline
 nose tip       & 0     & 36    &  2.5 \\ 
 left ear       & 7.75  & 16    & 19 \\ 
 right ear      & -7.75 & 16    & 19 \\ 
 left front paw & 5.5   & 20    & -8 \\
 right front paw & -5.5  & 20   & -8 \\
 left hind paw  & 13.5  & -8.5  & -8 \\
 right hind paw & -13.5 & -8.5  & -8 \\
 tail root      & 0     & -30   & -6 \\
 \hline
\end{tabular}
\label{tab:rigidMouseModel}
\end{table}

\subsection{Deep-Learning\\Deformable Mouse Model}

In a first approximation, real mouse's body part movements relative to the rigid model's theoretical body part positions can be treated like image co-ordinate measurement noise. However, this is not free of difficulties. For instance, the commonly used Gaussian noise model of imaged points would only be a very rough approximation of reality.

For this reason, in the attempt to model the real body's deformations, we follow a more elaborate approach. Estimated rigid body part image positions $\mathbf{\rigid{x}}_{t,i}$ appear in pairs with observed body part image points $\mathbf{\deformable{x}}_{t,i}$. Optionally, we deproject image points $\mathbf{x}$ to 3D object space points $\mathbf{X}$. Then the ill-posedness of deprojection has to be handled reasonably, e.g.~by the assumption of mouse body parts being localized on specific planes in 3D space that are determined by the mouse rigid body.

\begin{figure}[htbp]
  \includegraphics[width=\columnwidth]{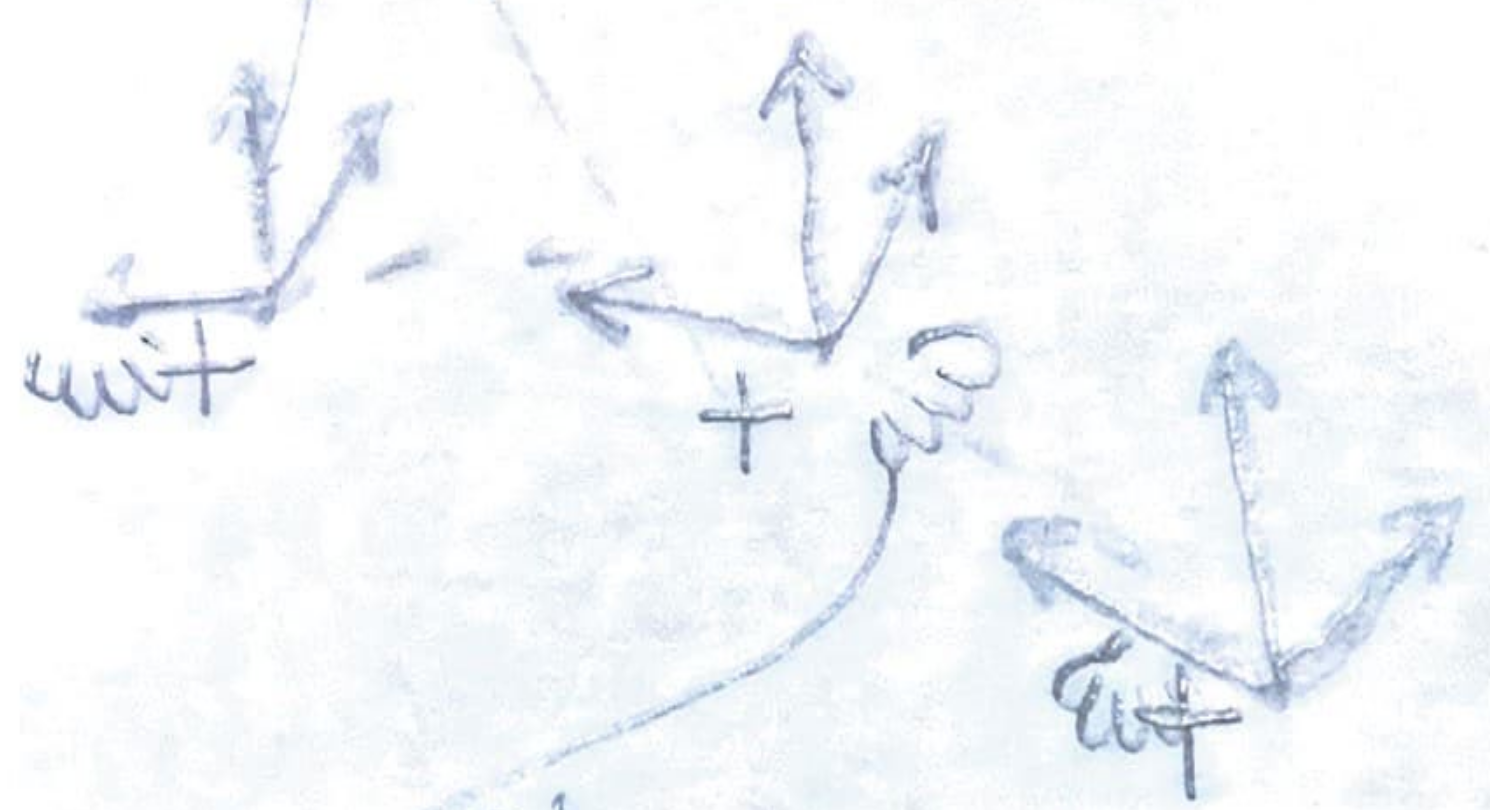}
  \caption{Tokens input to a deep net. The orthonormal vector triplets indicate the mouse model moving through space, plus ("+") signs indicate rigid body part, and paw "wiggles" indicate deforming body part.}
  \label{fig:transformerData}
\end{figure}

The pairs of rigid and deformable body parts can be considered the tokens
\begin{equation}
\label{Eq:token}
    \mathbf{T}_{t,i} = (\mathbf{\rigid{X}}_{t,i}, \mathbf{\deformable{X}}_{t,i})
\end{equation}
of a sequence over time.
The sequence
\begin{equation}
\label{Eq:sentence}
    \mathbf{V} = \begin{bmatrix}
        \mathbf{T}_{t-n} \cdots \mathbf{T}_{t} \cdots \mathbf{T}_{t+n} \end{bmatrix}'
\end{equation}
is equivalent to a sentence consisting of words with each tuple or co-ordinate vector being equivalent to a word (cf. Fig.~\ref{fig:transformerData}.

Therefore, we suggest to consider $\mathbf{V}$ the key (and query) tokens of a transformer deep net. They are directly considered as the non-contextual embeddings in the input layer.
This seems reasonable as word embeddings in natural language processing (NLP) are vectors corresponding to co-ordinates.
As opposed to NLP, we do not mask the last token, but part of the mid token of a sequence, namely $\mathbf{\deformable{X}}_{t,i}$. At training as well as test time it is predicted by the network. During training the deviation between predicted and observed co-ordinate is used to compute the loss.

At test time, we use the transformer net as an extension of observation equation Eq.~(\ref{Eq:observation})
\begin{equation}
    \label{Eq:extObservation}
    \mathbf{\deformable{x}}_{ti}^k = \mathcal{T}(\mathbf{P}_t^k \mathbf{X}_i)
\end{equation}
with transformer net $\mathcal{T}$ receiving the rigid model's body part image coordinate $\mathbf{\rigid{x}}$ as input producing deformable body part's image co-ordinates. They are compared with the observed image co-ordinates in the least squares adjustment (described in Section \ref{sec:lsqadj}) constraining the unknowns accordingly. Obviously, the extended observation Eq.~(\ref{Eq:extObservation}) describes the relation between mouse body and body part observation much more naturally than the original Eq.~(\ref{Eq:observation}).

\subsubsection{First Experiment Based on Simulated Data}

For the first deep-learning experiment we use simulated data. The simulated data is provided as tuples
\begin{equation}
    \label{Eq:simData}
    \mathcal{D}_t = \{t_t, \{i, \mathbf{\rigid{X}}_t,
    \mathbf{\deformable{X}}_t\}_{i=1}^{i=M}\}
\end{equation}
per time epoch $t$ with $M$ model parts as given in Table \ref{tab:rigidMouseModel}. The simulated data allow experiments with sequences of completely given data, as well as with missing data (e.g.~due to body part occlusions).

We are currently searching for an adequate transformer model and a way to implement it. For instance, a Pytorch model following \cite{b4} could be used \cite{b4}. Also Hugging face provides examples \cite{b2,b3}.

So far, conducted experiments indicate that LSTM networks may provide a more successful DL architecture than transformers.

\subsection{Least-Squares Adjustment}
\label{sec:lsqadj}

\subsubsection{Functional Model}

For the combined least-squares bundle adjustment all transformations are formulated as concatenated transformations from mouse model to camera $k$ to global (lock box) co-ordinates where a single vector of the unknown six trajectory parameters $\mathbf{p}_g$ resulting from transformations
\begin{equation}
    {\mathbf{H}_t}_m^g = {\mathbf{H}_g^k}^{-1} * {\mathbf{H}_t}^k
\end{equation}
is estimated at every time epoch $t$. Then incomplete mouse body part observations from all time epochs $t$ and from all three cameras $k$ support the mouse path estimation according to Eq.~(\ref{Eq:extObservation}) together with the smoothness constraints according to Eq.~(\ref{Eq:spline}).

The above derivation explains the functional model of linearized least-squares adjustment. Observations are the image co-ordinates of the mouse body parts in all three cameras, e.g. resulting from DeepLabCut mouse tracker \cite{mathis2018deeplabcut}. Unknowns are the three rotation and three translation parameters of the mouse model in global co-ordinates as a function of time $t$. An implementation of the adjustment is feasible with e.g.~the Ceres optimization software framework \cite{b6}. 

\subsubsection{Stochastic Model}

There are observation equations for body part image co-ordinates as given by Eqs.~(\ref{Eq:observation}) or (\ref{Eq:extObservation}) and the track constraint condition Eq.~(\ref{Eq:spline}) both providing residuals. 

Usually, image co-ordinates can be considered equally accurate and circularly Gaussian distributed. As here a difference is made between the technical geometric accuracy of body part identification and the assumption that body deformations can be modeled by a Gaussian-distributed image co-ordinate error, at least two Gaussians with extremely differing variances have to be considered.

The track constraint has to be weighted relative to the image co-ordinates stochastically described by the Gaussians. We do so by setting empirical weights in the cost function of the least-squares optimization (cf.~Ceres solver \cite{b6}).

\subsubsection{Approximations for Initialization}

What remains to be discussed is the initialization of the unknowns in order to allow linearized adjustment. This initial solution could stem from the implemented Kalman-Filter approach as described in \cite{Boon2024.07.29.605658}. However, it is also possible to initialize based on local observations according to Eq.~(\ref{Eq:observation}) only - as long as there are more then three observed body parts for a time epoch. Estimates for time epochs with less observations can be interpolated.

\section{Data}

\subsection{Simulated Data}

\begin{figure}[htbp]
  \centering\includegraphics[width=0.7 \columnwidth]{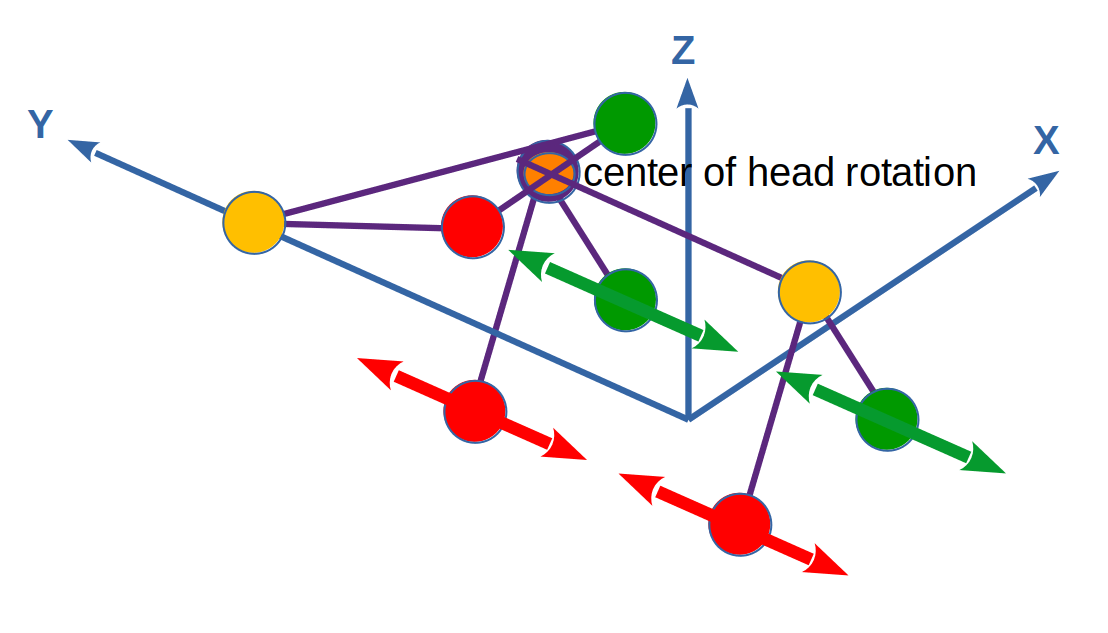}
  \caption{Simulated deformable mouse model}
  \label{fig:deformMouse}
\end{figure}

\begin{figure*}[!tb]
  \centering\includegraphics[width=\textwidth]{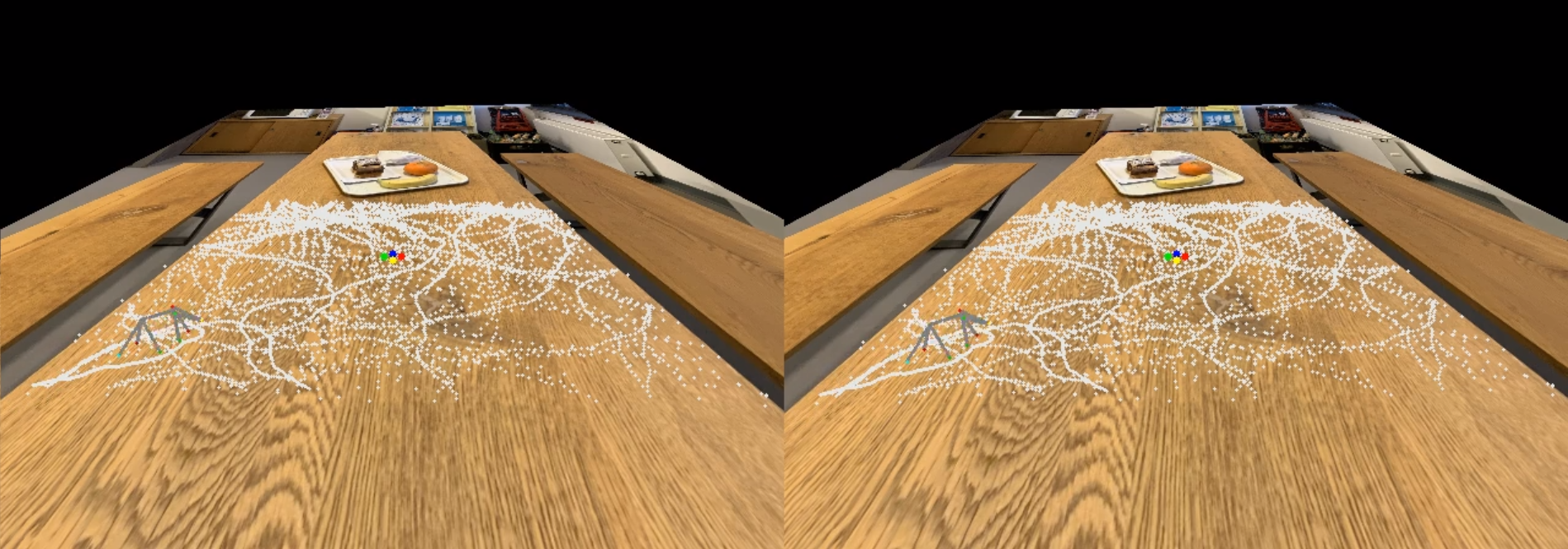}
  \caption{Stereo image pair of a simulated deformable mouse moving on a table. The track of the mouse model on the table plane is shown in white.}
  \label{fig:stereoSimDeformMouse}
\end{figure*}

The approach is tested with help of simulated data. For that purpose we let the rigid mouse model (cf.~Section \ref{sec:rigidMouse}) move along a 2D random curve \cite{b5} (random walk/Brownian motion ...) on a quadratic plane filling the field of view of cameras. During the model movement along the curve, the body parts deform plausibly (Fig.~\ref{fig:deformMouse}). Paws move according to pace (which may rather suit horses than mice): While right front and left hind paw are fixed to the ground, left front paw and right hind paw move forwards twice as fast as the overall mouse body. Meanwhile the rigid head triangle consisting of nose tip and ears rotates linearly back and force in three limited angle intervals around the mid point of the line connecting the two ears. Fig.~\ref{fig:stereoSimDeformMouse} shows a stereo image view of a simulated mouse moving on a table scenery.

\subsection{Real Data}

\begin{figure*}[!tb]
    \centering
    \begin{subfigure}[b]{0.475\textwidth} 
        \centering
        \includegraphics[width=\textwidth]{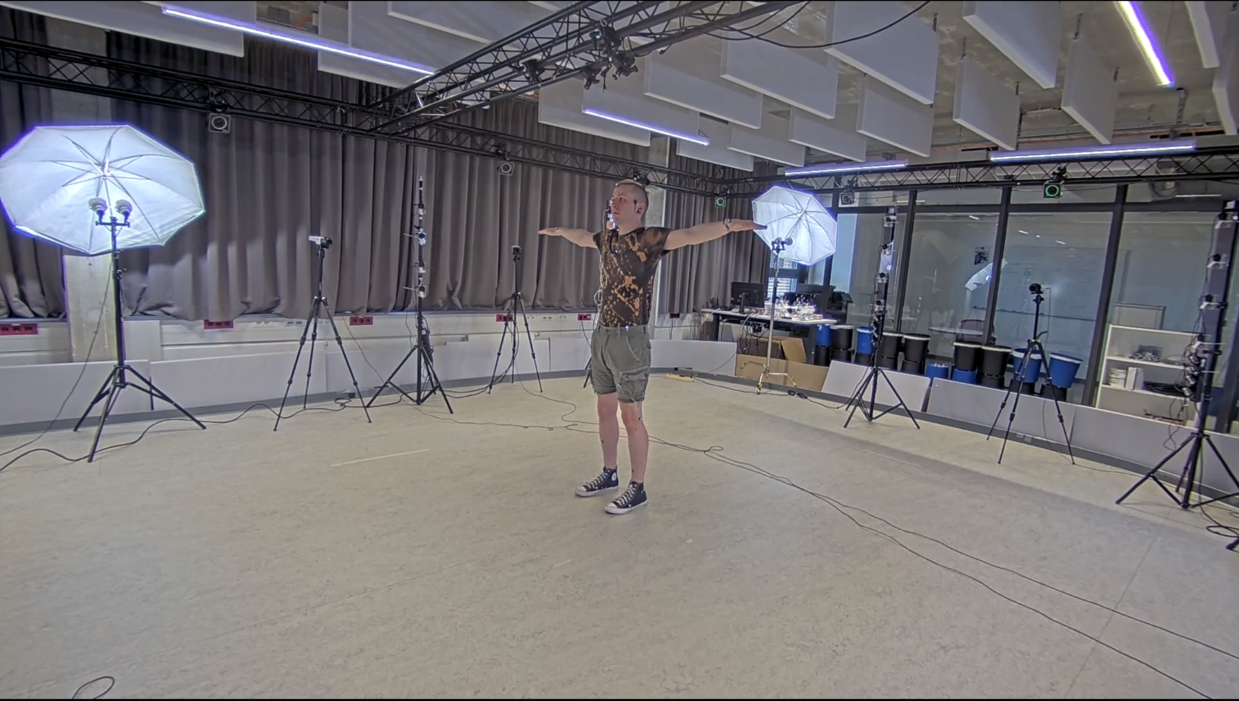}
        \caption[]%
        {{\small view 0}}    
        \label{fig:view0}
    \end{subfigure}
    \hfill
    \begin{subfigure}[b]{0.475\textwidth}  
        \centering 
        \includegraphics[width=\textwidth]{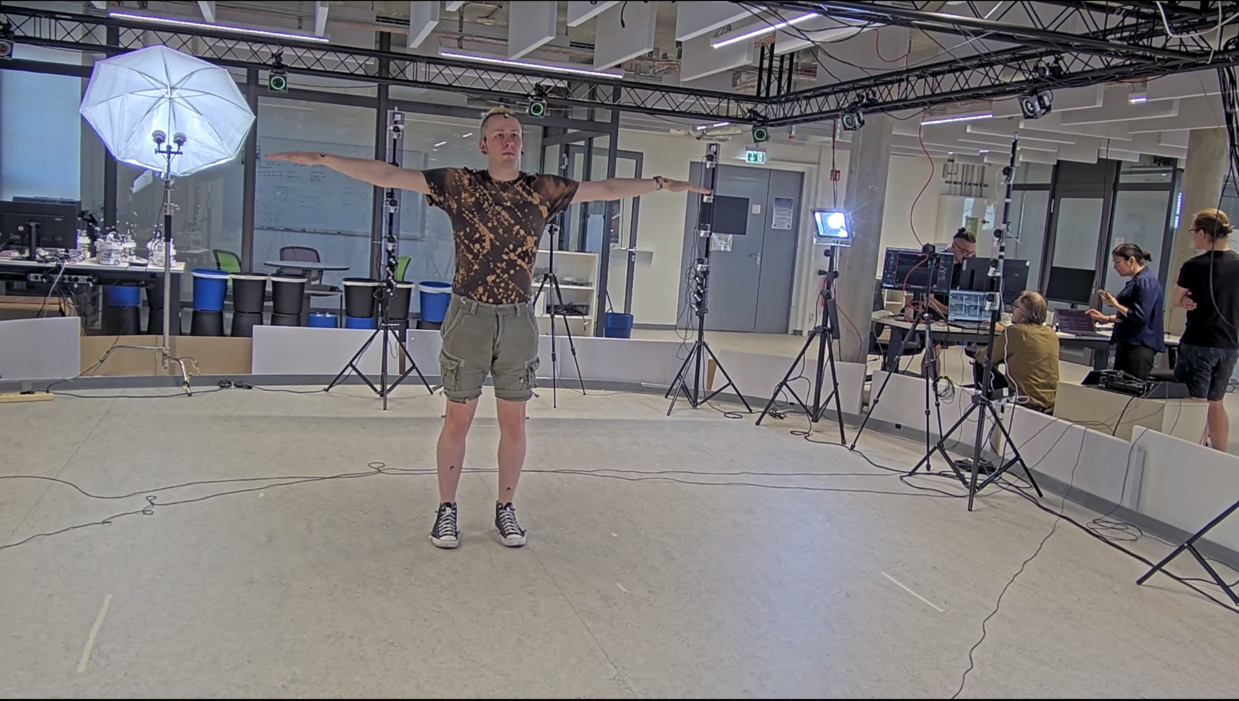}
        \caption[]%
        {{\small view 1}}    
        \label{fig:view1}
    \end{subfigure}
    \vskip\baselineskip
    \begin{subfigure}[b]{0.475\textwidth}   
        \centering 
        \includegraphics[width=\textwidth]{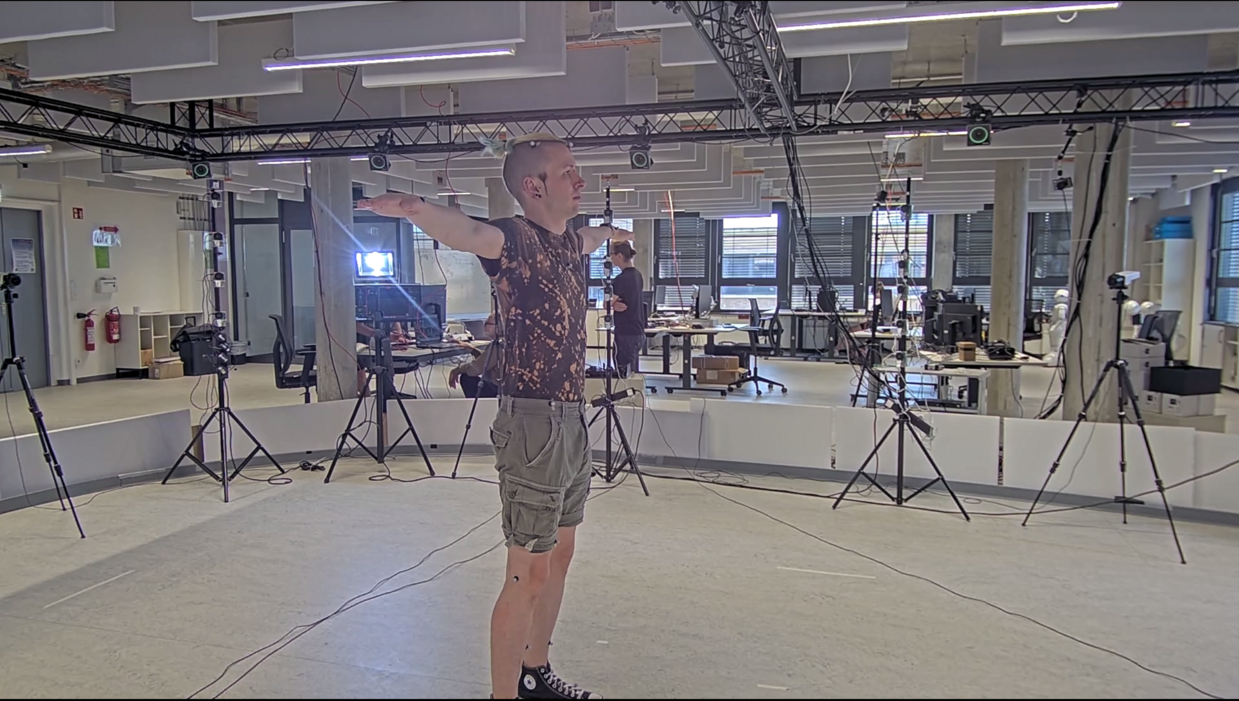}
        \caption[]%
        {{\small view 2}}    
        \label{fig:view2}
    \end{subfigure}
    \hfill
    \begin{subfigure}[b]{0.475\textwidth}   
        \centering 
        \includegraphics[width=\textwidth]{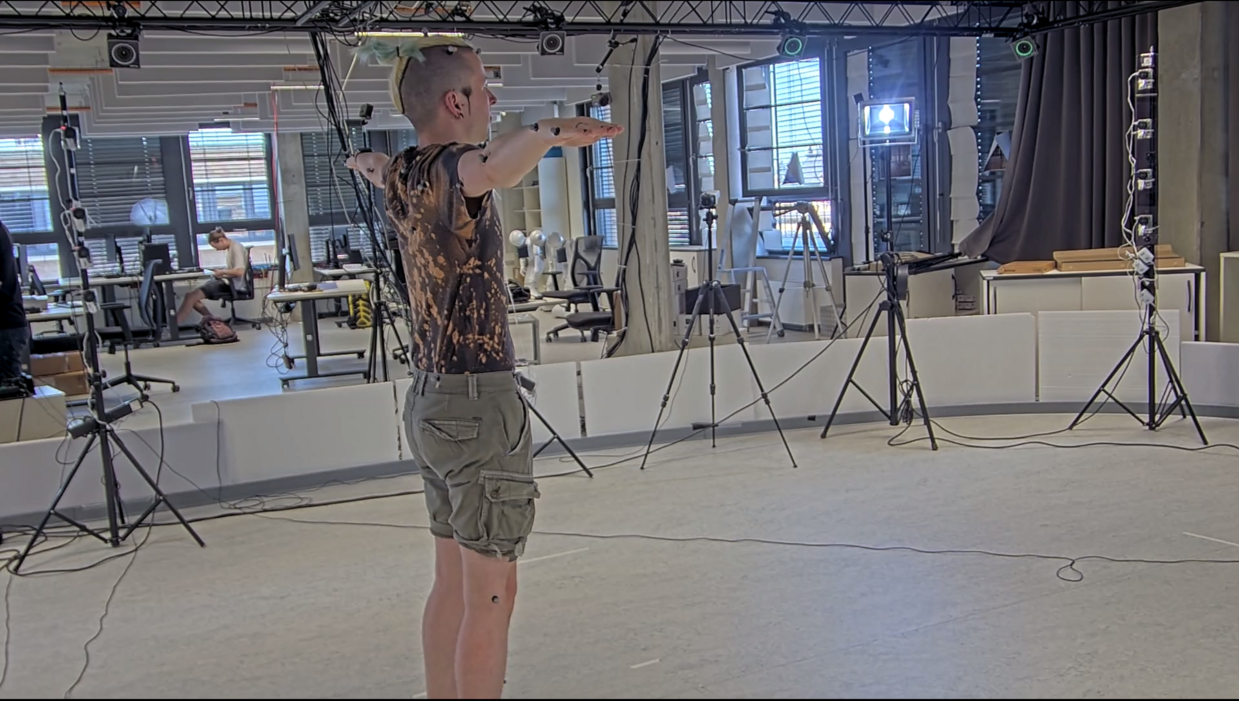}
        \caption[]%
        {{\small view 3}}    
        \label{fig:view3}
    \end{subfigure}
    \caption[]
    {Multi-view motion capturing with several cameras at time epoch $t$} 
    \label{fig:motionCapture}
\end{figure*}

The approach cannot only be applied to the previously described mouse lock box videos, but also to e.g.~observations of humans. Fig.~\ref{fig:motionCapture} shows a human executing movements in a motion capture lab. Besides the IR cameras for tracking spherical retro-reflective markers on the clothing, in the setup shown conventional cameras of two types (Raspberry Pi cameras and Mangold Video Observation Lab cameras) are used to record movements of the proband. Depending on how many cameras are available the suggested approach can be of importance. In particular, with a small number of video cameras in presence of occluding objects, a body model including body-part motions and the suggested motion-track constraint can be helpful.

Equivalent to the rigid model suggested for mouse tracking could be an articulate model such as the SMPL model \cite{bogo2016keep}. Then the dynamic part of the model would only be responsible for deviations from the results of the articulate model's components detections.

\section{Outlook to Behavior Analysis}

Behavior can be characterized by movements, e.g.~movements indicating that an individual manipulates a mechanism present in its environment. So movements are suitable to classify behavior. 

Here, movements are extracted from video in order to be used for behavior classification allowing to identify interactions with lock-box parts. Compared to the input data, the output data generated with help of the suggested approach is both more complete and more geometrically accurate which is why behavior analysis based on our model's output is more promising than direct analysis of the input data.

Implicit prerequisite of the anticipated improvement of behavior analysis is that the mix of deep-learned body deformations and linearized least-squares adjustment leads to more correct body-part motion estimates. We will demonstrate the improvement of estimates by comparing behavior analysis based on original data with behavior analysis based on the method's output data.

In a supervised downstream application the trained transformer net could be used for behavior classification in a similar manner as a language model pretrained by sentence completion is used for sentiment classification (cf.~\cite{b1}).

\section{Conclusions}

In this conceptual work, an approach for the analysis of behavior based on video observations is suggested that explicitly estimates 3D body part positions as intermediate results before deriving final conclusions w.r.t.~e.g.~behavior classes. For this purpose, a motion-track constraint is imposed on body-part movements as a condition in a multi-view least-squares bundle adjustment. Secondly, deep-learned body-part movements are used to model movements relative to a rigid body model. In future work, we will present the approach including experimental results in more detail.

The suggested approach competes with approaches estimating behavior directly from observed video frames without intermediate 3D observations. We will also attend to the development of such approaches.

\section{Acknowledgments}

Funded by the Deutsche Forschungsgemeinschaft (DFG, German Research Foundation) under Germany’s Excellence Strategy – EXC 2002/1 “Science of Intelligence” – project number 390523135.

\bibliography{references.bib}
\end{document}